\documentclass[11pt]{article}

\usepackage[preprint]{acl}
\usepackage{booktabs}
\usepackage{textcomp}
\usepackage{listings}
\usepackage{xcolor}
\usepackage{multirow}
\usepackage{booktabs}

\lstdefinestyle{promptstyle}{
  basicstyle=\ttfamily\scriptsize,
  frame=single,
  breaklines=true,
  breakatwhitespace=false,
  columns=fullflexible,
  keepspaces=true,
  showstringspaces=false,
  tabsize=2,
  captionpos=b
}

\usepackage{times}
\usepackage{latexsym}
\usepackage{amssymb}
\usepackage{amsmath}

\usepackage[T1]{fontenc}

\usepackage[utf8]{inputenc}

\usepackage{microtype}

\usepackage{inconsolata}

\usepackage{graphicx}
\usepackage{xcolor}

\title{Faithful Summarisation under Disagreement via Belief-Level Aggregation}
\author{
  Favour Yahdii Aghaebe\thanks{School of Computer Science}\thanks{Healthy Lifespan Institute},
  Tanefa Apekey\thanks{Sheffield Centre for Health and Related Research},
  Elizabeth Williams\thanks{Department of Oncology and Metabolism}\footnotemark[2],
  Nafise Sadat Moosavi\footnotemark[1] 
  \\
  University of Sheffield, UK \\
  \texttt{\{fyaghaebe1, t.apekey, e.a.williams, n.s.moosavi\}@sheffield.ac.uk}
}

\begin{document}
\maketitle
\begin{abstract}
Opinion and multi-document summarisation often involve genuinely conflicting viewpoints, yet many existing approaches, particularly LLM-based systems, implicitly smooth disagreement and over-represent majority opinions. This limits the faithfulness of generated summaries in opinion-heavy settings.
We introduce a disagreement-aware synthesis pipeline that separates belief-level aggregation from language generation. Documents are first represented as structured belief sets and aggregated using distance-based belief merging operators that explicitly model conflict. Large language models are then used only to realise the aggregated beliefs as natural language summaries.
We evaluate the approach across multiple model families and scales, comparing it to methods that perform explicit aggregation during generation. Our results show that while sufficiently large models can match belief-level aggregation when aggregation is handled at generation time, this behaviour is not stable across architectures or capacities. In contrast, belief-level aggregation combined with simple prompting yields consistently strong disagreement-aware performance across models, while maintaining fluent and grounded summaries. 
\end{abstract}

\section{Introduction}
Summarisation is often treated as a task of condensation: distilling multiple inputs into shorter, coherent representations. This framing is effective when source documents are broadly consistent. However, it becomes problematic when inputs encode incompatible evidence or opinions. In such cases, coherence is not an inherent property of the data, but an artefact imposed by the summarisation process itself. The result is a subtle but consequential failure mode: disagreement is collapsed, minority perspectives are attenuated, and summaries project a consensus that does not exist.

This mismatch is especially apparent in multi-document and opinion summarisation. Opinions are subjective expressions of attitudes or evaluations toward entities or their aspects \citep{Maharani2017AspectOpinionSurvey, Kim2013OpinionSummarization}, and collections of opinions frequently exhibit genuine and irreducible disagreement. Opinion summarisation aims to condense such collections into a form that is accessible to end users \citep{liu2020opinion, Kim2013OpinionSummarization, LOPEZCONDORI2017124}. Yet a substantial body of work shows that existing approaches, both extractive and abstractive, tend to privilege majority viewpoints, producing summaries that smooth over disagreement and marginalise minority or dissenting opinions \citep{suhara-etal-2020-opiniondigest, li2024rationalebasedopinionsummarization, huang-etal-2024-bias}. As a result, summaries may appear balanced while failing to reflect the true structure of the underlying opinions.

Recent large language model–based approaches inherit and often amplify this tendency. While LLMs generate fluent and well-structured summaries \citep{zhang2025comprehensivesurveyprocessorientedautomatic}, empirical studies show that they are prone to drift, overgeneralisation, and the introduction of unsupported inferences \citep{peters2025generalizationbiaslargelanguage, aghaebe2025llmsageassessingdemographic}. In the presence of conflicting inputs, these behaviours manifest as implicit conflict resolution: disagreement is averaged away, majority positions dominate, and summaries convey a spurious sense of agreement \cite{suhara-etal-2020-opiniondigest, huang-etal-2024-bias, peters2025generalizationbiaslargelanguage}. Crucially, these failures arise not from missing information, but from how information is aggregated.

In this work, we examine summarisation under disagreement through the lens of \emph{where aggregation happens}. Most LLM-based approaches implicitly assume that conflicting information can be reconciled \emph{during} generation: the model is given multiple reviews and expected to internally decide which viewpoints to retain, how to balance them, and how to express the result. Our experiments show that this assumption holds only for a subset of models. For some strong models, such as GPT-5 and Qwen-14B, generation-time fusion can yield reasonable disagreement-aware summaries. However, the same setups degrade sharply for smaller or differently trained models, revealing aggregation behaviour that is brittle, model-dependent, and difficult to control. 

Motivated by this instability, we instead treat multi-review synthesis as an \emph{explicit aggregation} problem. We draw on \emph{belief merging}, a formal framework from logic and philosophy that studies how multiple, potentially inconsistent belief sets can be combined into a collective representation \citep{bloch_fusion_2001,konieczny2002merging}. Distance-based belief merging operators explicitly model conflict and select compromise belief states that minimise global disagreement, rather than collapsing opinions through majority voting or ad hoc heuristics. This reframing separates two challenges that are typically entangled in end-to-end summarisation: deciding \emph{what} the collective stance should be, and deciding \emph{how} to express it. In our pipeline, aggregation is handled upstream via belief merging over structured aspect-level opinions, while the LLM is used only for realisation through a lightweight synthesis prompt.
Our results show that making aggregation explicit fundamentally changes the role of the language model. Once disagreement is resolved at the belief level, even simple prompting yields stable and competitive performance across model families and scales. In contrast to generation-level fusion, belief merging substantially reduces sensitivity to model capacity and architecture: weaker models achieve disagreement-aware scores comparable to stronger ones, without specialised synthesis prompts. We compare this explicit-aggregation approach against QA-prompting \citep{sinha-2025-qa}, which delegates aggregation to the model via intermediate question answering, and against an adaptation of Fusion-of-N \citep{khairi2025makingtakingbestn}, a sophisticated generation-level fusion strategy. While Fusion-of-N can perform well for certain models, its effectiveness does not consistently transfer across scales.
Taken together, our findings suggest that the core challenge in summarisation under disagreement is not linguistic generation, but the stability of aggregation. When aggregation remains implicit, performance varies widely across models; when aggregation is made explicit, disagreement-aware synthesis becomes markedly more robust. This indicates that structured aggregation can substitute for model capacity in this setting, enabling faithful synthesis of conflicting opinions without relying on increasingly powerful generators.

\section{Related Work}
\subsection{Summarisation under Disagreement}
Disagreement arises in many multi-document summarisation settings involving conflicting evidence or conclusions. Opinion summarisation provides a particularly clear and well-studied instance of this broader challenge.
Earlier work on opinion summarisation predominantly relied on \emph{extractive} techniques, which aim to identify and select salient sentences or phrases directly from a collection of opinions and concatenate them into a summary \citep{10.1145/1014052.1014073, nenkova-etal-2011-automatic}. A representative example is the feature-based opinion summarisation framework of \citet{10.1145/1014052.1014073}, which extracts representative sentences for product features based on sentiment polarity and frequency. While extractive methods are relatively simple and preserve faithfulness to the source text, they suffer from well-documented limitations, including redundancy,  abrupt topic shifts, and poor fluency when sentences from different authors are juxtaposed \citep{10.1561/1500000011, nenkova-etal-2011-automatic}.

To address these limitations, \emph{abstractive} opinion summarisation methods have been proposed. Rather than copying sentences verbatim, abstractive approaches aim to generate novel sentences that capture and fuse the underlying ideas, sentiments, and aspects expressed across multiple opinions \citep{LOPEZCONDORI2017124}. By operating at a more semantic level, these methods can reduce redundancy, improve coherence, and produce more readable and human-like summaries. However, abstractive opinion summarisation introduces its own challenges, such as content selection, sentiment faithfulness, and the risk of hallucinating unsupported opinions.

A central challenge shared by extractive and abstractive opinion summarisation is the \emph{opinion smoothing problem}, where fine-grained and diverse opinions are overly aggregated into a single sentiment \cite{amplayo-lapata-2021-informative}. This can obscure meaningful variation in user feedback, especially when strong minority opinions are present. Closely related is the tension between representing \emph{majority} versus \emph{minority} opinions. Many summarisation systems implicitly prioritise majority viewpoints by relying on frequency-based signals, which can marginalise less frequent but potentially important perspectives \citep{huang-etal-2025-refer}. 

Advances in neural sequence-to-sequence models, and more recently large language models, have significantly improved abstractive opinion summarisation. Pretrained LLMs demonstrate strong zero-shot and few-shot capabilities due to their broad world knowledge and powerful language generation abilities \citep{zhao2025surveylargelanguagemodels}. When applied to opinion summarisation, these models can generate fluent, coherent, and aspect-aware summaries, often improving readability over earlier neural approaches. However, while LLMs substantially improve generation quality, they do not necessarily resolve longstanding challenges in opinion summarisation, including controllability, faithfulness to source opinions, and potential amplification of majority biases \citep{zhao2025surveylargelanguagemodels}. Consequently, recent work has explored mechanisms that complement generative models with explicit structures for opinion aggregation, diversity preservation, or intermediate grounding, to mitigate opinion smoothing and improve faithfulness under disagreement \citep{amplayo-lapata-2021-informative, suhara-etal-2020-opiniondigest, li2024rationalebasedopinionsummarization, huang-etal-2025-refer}.

Beyond opinion summarisation, related challenges have also been identified in multi-document summarisation of conflicting evidence, such as scientific and systematic review synthesis, where models may drift toward spurious consensus rather than faithfully aggregating incompatible claims \cite{10.1162/tacl_a_00687,peters2025generalizationbiaslargelanguage}. Our work aligns with this broader line of research in treating aggregation under disagreement as an explicit component of summarisation.

\subsection{Opinion Diversity and Bias}
Initial studies on fairness and bias in summarisation primarily focused on ensuring that summaries accurately reflected the diverse social or demographic attributes present in the source documents \citep{dashgenerated2019, keswani2021dialectdiversitytextsummarization, olabisi-etal-2022-analyzing, Shandilya2020FairnessFW, zhang-etal-2024-fair}. These approaches emphasised representational fairness, aiming to prevent the systematic exclusion or misrepresentation of particular social groups.

Recent work has, in addition, begun to examine fairness in opinion summarisation from the perspective of opinion coverage rather than demographic representation. \citet{huang-etal-2023-examining} highlight that opinion summarisation systems may suppress minority viewpoints even when producing semantically similar summaries. However, their stance-based evaluation does not explicitly address how conflicting opinions should be aggregated in a principled manner. 
Subsequent approaches have explored structured prompting strategies to mitigate majority bias by introducing intermediate representations. \citet{huang-etal-2025-refer} propose a frequency-framed prompting method for aspect-based summarisation, in which models are prompted to extract opinion frequencies prior to generation. While this strategy can improve coverage of minority opinions in larger and proprietary models, its effectiveness is less consistent in smaller models, which may struggle to reliably perform the intermediate frequency extraction step. Similarly, \citet{sinha-2025-qa} introduce a Question Answering–based approach, where summaries are generated from answers to prompted questions. In opinion summarisation settings, however, such methods may still privilege dominant viewpoints when the elicited answers primarily reflect majority perspectives. 
Related work has also promoted opinion diversity through explicit control over input selection rather than aggregation. In particular, \citet{jiang-etal-2023-large} propose a supervised framework for large-scale multi-perspective opinion summarisation that constructs diverse review subsets via sentiment-aware sampling and information valuation. While effective at encouraging exposure to heterogeneous viewpoints during training, diversity is primarily enforced at the level of review selection. Conflicting opinions are handled implicitly by conditioning the model on sentiment-controlled subsets, rather than through an explicit mechanism for modelling disagreement or determining how minority views should influence the final summary.

\subsection{Belief Merging and Judgement Aggregation}
Belief merging originates in knowledge representation and formal epistemology, where it addresses the problem of combining multiple, potentially inconsistent belief bases into a single collective belief state while respecting logical consistency and fairness principles \citep{konieczny2002merging}. The central concern of belief merging is how to aggregate information from multiple sources when conflicts cannot be resolved by simple conjunction or voting. To this end, belief merging frameworks explicitly model disagreement and define operators that seek compromise solutions, minimising global inconsistency across sources.
Belief merging has been studied as a principled alternative to majority-based schemes, which may yield inconsistent or unfair outcomes when agents hold incompatible beliefs. A wide range of merging operators have been proposed, including distance-based approaches that characterise collective beliefs in terms of minimal revision from individual belief bases \citep{konieczny2002merging}. These operators provide formal guarantees about consistency and rationality, making belief merging a well-established framework for reasoning under disagreement.

Although belief merging has traditionally been formulated in propositional logic, its underlying perspective has informed applied settings beyond formal epistemology. For example, \citet{kareem_application_2017} adapt belief merging operators within a machine learning pipeline for medical decision support, demonstrating how principled belief aggregation can be integrated with data-driven models. Similarly, \citet{wilie2024beliefrevisionadaptabilitylarge} utilise belief revision, a closely related phenomenon, to evaluate LLM reasoning ability. 

Taken together, this line of work underscores the importance of explicitly accounting for diversity and disagreement in summarisation, while also revealing the limitations of heuristic or prompt-based approaches to conflict handling. Building on these insights, we frame multi-document summarisation under disagreement as an aggregation problem rather than a purely generative one. Although belief merging has seen limited use in natural language summarisation, its explicit treatment of conflict offers a principled way to structure disagreement prior to generation.

\section{Method}
\label{sec:method}
We propose a synthesis pipeline \textsc{AccSynth} that separates aggregation under disagreement from natural language generation. The pipeline consists of three stages: (1) extraction of structured opinion-level information from source documents, (2) aggregation of this information via belief merging, and (3) LLM-based synthesis of the merged beliefs into a natural language summary.

\paragraph{Structured Opinion Representation.} 
Given a collection of documents $\mathcal{D} = \{d_1, \dots, d_n\}$, we extract structured representations that expose disagreement. In the opinion summarisation setting, each document corresponds to a single reviewer and is decomposed into a set of aspect-level opinions of the form $\langle \text{aspect}, \text{polarity} \rangle$. Each document $d_i$ is thus mapped to a belief base $K_i$, defined as a finite set of propositional statements encoding the opinions expressed in that document. This representation abstracts away from surface realisation while preserving the informational content necessary for aggregation.

\paragraph{Belief Merging for Aggregation under Disagreement.}

\begin{figure}[t]  
    \centering
    \includegraphics[width=0.38\textwidth]{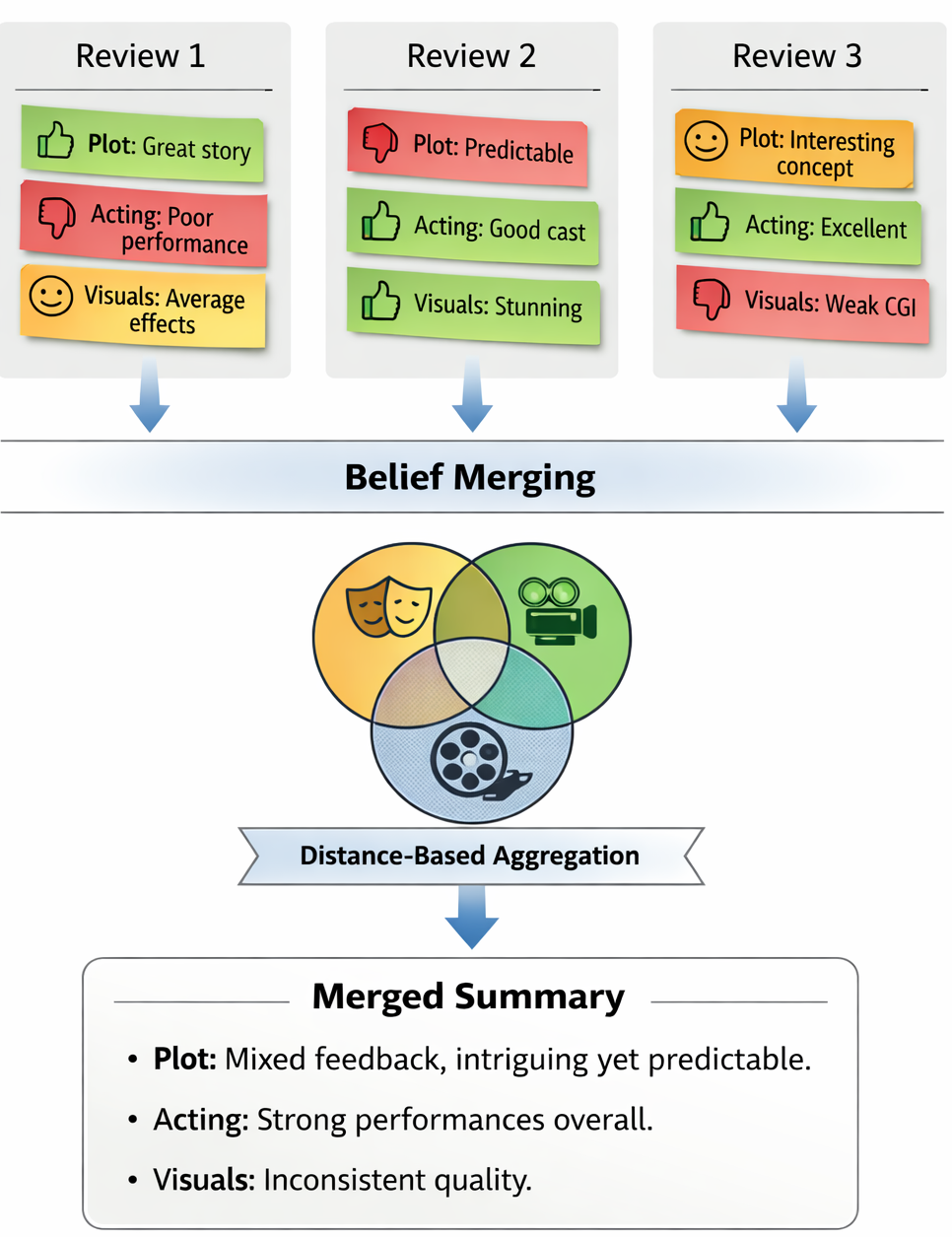} 
    \vspace{-0.2cm}
    \caption{Belief-based opinion aggregation. Aspect-level opinions from multiple reviews are aggregated via belief merging before summarisation.}
    \label{fig:bm}  
\end{figure}

Let $\mathcal{K} = \{K_1, \dots, K_n\}$ denote the profile of belief bases extracted from the source documents, where each $K_i$ encodes the aspect-level opinions expressed in document $d_i$. The goal of belief merging is to compute a cohesive belief base $K^{\ast}$ that represents the collective information in $\mathcal{K}$, even when the individual belief bases are mutually inconsistent. Figure~\ref{fig:bm} illustrates this aggregation process in the context of summarisation.

We adopt \emph{distance-based belief merging} \footnote{We provide more details in Appendix \ref{app:belief-merging}}, which formulates aggregation as an optimisation problem over possible interpretations. In our setting, an interpretation corresponds to a complete assignment of opinion polarities to aspects. Let $\Omega$ denote the set of such interpretations, and let $d(\omega, K_i)$ measure the extent to which an interpretation $\omega \in \Omega$ disagrees with the belief base $K_i$.

Given an aggregation function $f$, the total distance between an interpretation $\omega$ and the belief base $\mathcal{K}$ is defined as:
\[
D(\omega, \mathcal{K}) =
f\big(d(\omega, K_1), \ldots, d(\omega, K_n)\big).
\]

The distance-based belief merging operator $\Delta^{d,f}$ then selects the interpretation(s) that minimise this total distance while satisfying any predefined integrity constraints $\mu$:
\[
\mathrm{Mod}(\Delta^{d,f}(\mathcal{K}, \mu)) =
\arg\min_{\omega \models \mu} D(\omega, \mathcal{K}).
\]

Intuitively, this procedure selects the global description of the item that disagrees as little as possible with the individual sources taken together. This yields a compromise belief state that balances conflicting evaluations across aspects, rather than collapsing disagreement through majority voting or arbitrary tie-breaking. The resulting merged belief base $K^{\ast}$ serves as the input to the subsequent summarisation stage.

\paragraph{LLM-Based Summarisation from Merged Beliefs.}
Belief merging produces a quantitative belief base $K^{\ast}$ rather than a natural language summary. 
We therefore treat generation purely as a realisation step: the language model is conditioned on the merged beliefs and is not responsible for resolving disagreement or selecting viewpoints.
All aggregation decisions are made upstream by the belief merging operator, and the model operates only over the aggregated belief state. This separation makes disagreement handling explicit and model-independent.

\section{Experimental Design}
\subsection{Task Definition}
The target output for each instance is a rotten-tomato in-house critic meta-review. Given a set of movie critic reviews for the same item, models are tasked with generating a summary that accurately represents the opinions expressed across individual reviews.

\subsection{Dataset}
We use the Rotten Tomatoes movie review dataset \citep{leone_rotten_2020}, which comprises approximately 17,000 critic reviews together with corresponding critic meta-reviews written by Rotten Tomatoes staff. Around 8,000 movies in the dataset have an associated meta-review, which serves as the reference summary in our experiments. Following standard practice, we partition the data into Train, Dev, and Test splits. As our approach does not involve model fine-tuning, the Train and Dev splits are used only for prompt verification and methodological validation, while all reported results are obtained exclusively on the held-out Test split. Dataset statistics are reported in Table~\ref{tab:dataset}.

A substantial portion of the dataset exhibits genuine opinion disagreement. Using an LLM-based classifier (Gemini~2.5~Pro; \citealp{geminiteam2025geminifamilyhighlycapable}), we analyse the distribution of disagreement across key movie aspects and find that approximately 40\% of movies contain explicitly opposing evaluations on at least one aspect, with 366 movies showing particularly pronounced disagreement (i.e., more than 50\% of critics disagreeing on an aspect). This prevalence of conflict motivates our focus on synthesis under disagreement and underscores the limitations of approaches that implicitly smooth opinions during generation.

\subsection{Method Instantiation}
\paragraph{Opinion Extraction.}
We begin by extracting each review’s sentiment toward important movie-related aspects, which were identified using the Latent Dirichlet Allocation (LDA) topic modelling \citep{blei2003latent} algorithm (Figure~\ref{fig:topic_model}). Based on this analysis, we select a fixed set of aspects: \emph{Plot}, \emph{Themes}, \emph{Script}, \emph{Performances}, \emph{Visuals}, \emph{Pacing}, and \emph{Overall success}. These align with well-established expectations of movie review content \citep{filmdiscourse, simonton_cinematic_2009}. For each review, we extract aspect-level opinions of the form $\langle \text{aspect}, \text{polarity} \rangle$ using the Qwen/Qwen2.5-72B-Instruct model \citep{qwen3}. 

\paragraph{Belief Merging Configuration.}
In our experiments, the distance-based merging is operationalised using the L1 (Manhattan) distance, which corresponds to the expected Hamming disagreement between interpretations and belief bases. In instances where more than one interpretation has the lowest distance, aggregation is performed by averaging across selected interpretations, yielding a merged belief base $K^{\ast}$ that represents a probabilistic compromise across conflicting opinions.

\paragraph{LLM-Based Summarisation from Merged Beliefs}
To realise $K^{\ast}$ as text, we perform a two-stage LLM-based synthesis process that separates belief verbalisation from summary generation.
First, for each aspect-level belief in $K^{\ast}$, we apply a world conversion prompt that maps a single merged belief, such as a positive, negative, or mixed evaluation of an aspect, into a short, critic-style natural-language statement. This step performs no aggregation or inference: the model is not asked to compare reviews, resolve disagreement, or decide which opinions to include. It simply verbalises an already-determined belief state, yielding a one-to-one correspondence between merged beliefs and their linguistic realisations. Second, the resulting set of belief statements is summarised into a coherent paragraph using a lightweight summarisation prompt. Because all aggregation decisions are fixed upstream, this final generation step focuses only on fluency and coherence, rather than resolving conflicting viewpoints. This design keeps disagreement handling explicit and upstream of generation, while allowing the use of simple, uniform prompting across models.

We also experiment with an alternative instantiation in which the verbalised beliefs are combined using an adaptation of Fusion-of-N at the synthesis stage, rather than direct summarisation. This variant tests whether more complex, generation-level fusion yields additional gains once belief-level aggregation is already explicit \footnote{All prompts are included in Appendix~\ref{appendix:prompts}}.

\subsection{Comparison Methods}
We compare our approach against two recent baselines that reflect alternative ways of handling multi-review synthesis, differing in how and where aggregation is performed.
First, we include the QA-based prompting method of \citet{sinha-2025-qa}, which reformulates opinion summarisation as a sequence of question–answering steps over the review set. This approach relies on structured prompts to surface salient information, with aggregation handled implicitly by the language model, making it a representative baseline for prompt-driven aggregation without explicit intermediate representations.
Second, we consider an adaptation of Fusion-of-N \citep{khairi2025makingtakingbestn}, a recent general-purpose method for synthesising information across multiple candidate inputs in LLM generation. Fusion-of-N provides a relevant comparison as a generation-level aggregation strategy that combines information from multiple sources rather than selecting a single input. To align it with our setting, we apply Fusion-of-N directly to the full set of critic reviews, treating each review as a candidate to be fused into a single summary. This baseline allows us to assess the extent to which multi-candidate generation alone can capture divergent opinions.

\subsection{Model Selection}
We evaluate models at two representative scales—small (1B–9B) and medium (>10B)—drawing from the Qwen, LLaMA, and OpenAI model families.\footnote{Further details, including computational expenses and model hyperparameters, are included in Appendix \ref{appendix:taskdetails}.}

\paragraph{Qwen.}
We evaluate two open-source autoregressive transformer models from the Qwen family (Qwen3-4B-Instruct-2507 and Qwen3-14B; \citealp{qwen3}). These models are selected for their instruction-following capabilities and long-context reasoning performance.

\paragraph{LLaMA.}
From the LLaMA family, we evaluate Llama-3.2-3B-Instruct and Llama-3.1-8B-Instruct \citep{grattafiori2024llama3herdmodels}.

\paragraph{GPT.}
We include a large proprietary model accessed via the OpenAI API (GPT-5; \citealp{openai2024gpt4technicalreport}) as a representative commercial system.

\subsection{Evaluation Metrics}
\label{sec:evaluation}
We conduct a supervised evaluation using the critic meta-review as the reference summary. Because our task focuses on \emph{faithful synthesis under disagreement}, rather than surface-level similarity alone, we combine reference-based scoring metrics with criteria-based LLM evaluation. These metrics provide complementary signals, capturing semantic alignment, fluency, and the faithful representation of conflicting opinions.

\subsection{Reference-Based Scoring Metrics}
We report BERTScore \citep{DBLP:journals/corr/abs-1904-09675} and BARTScore \citep{NEURIPS2021_e4d2b6e6} as reference-based metrics that measure semantic similarity and fluency relative to the gold critic meta-reviews. BERTScore evaluates similarity using contextual embeddings, while BARTScore computes the conditional likelihood of the reference given the generated summary. These metrics capture overall content alignment and readability.

\begin{table*}[!t]
\centering
\scriptsize
\setlength{\tabcolsep}{4pt}
\begin{tabular}{ll|ccccc|cc}
\toprule
\textbf{Method} & \textbf{Model} &
\multicolumn{4}{c}{\textbf{Disagreement-Aware (GEval)}} &
\multicolumn{1}{c}{\textbf{Groundedness (GEval)}} &
\multicolumn{2}{c}{\textbf{Reference-Based}} \\
\cmidrule(lr){3-6} \cmidrule(lr){7-7} \cmidrule(lr){8-9}
& &
\textbf{Cov} & \textbf{Pol} & \textbf{Ctr} & \textbf{Prev} &
\textbf{Grd} &
\textbf{BERT} & \textbf{BART} \\
\midrule

\multirow{5}{*}{Fusion-of-N}
& GPT-5        & 7.03 & 7.06 & 6.66 & 6.83 & 8.11 & 0.84 & \bfseries -3.32 \\
& LLaMA-3B     & 1.45 & 1.45 & 1.18 & 1.31 & 2.16 & 0.83 & -4.61 \\
& LLaMA-8B     & 3.84 & 3.81 & 3.25 & 3.52 & 4.96 & 0.80 & -4.55 \\
& Qwen3-4B     & 7.09 & 7.12 & 6.68 & 6.87 & 8.19 & 0.85 & -3.40 \\
& Qwen3-14B    & 7.17 & 7.30 & 6.36 & 6.67 & \bfseries 8.73 & 0.85 & -3.52 \\

\cmidrule(lr){1-9}

\multirow{4}{*}{QA Prompting}
& LLaMA-3B     & 4.30 & 4.45 & 1.95 & 3.08 & 2.72 & 0.86 & -3.52 \\
& LLaMA-8B     & 4.99 & 5.09 & 2.13 & 3.05 & 2.46 & 0.82 & -3.44 \\
& Qwen3-4B     & 5.30 & 5.81 & 2.78 & 4.29 & 3.85 & \bfseries 0.87 & -3.41 \\
& Qwen3-14B    & 5.04 & 5.49 & 2.66 & 3.99 & 4.00 & 0.86 & -3.51 \\

\cmidrule(lr){1-9}

\multirow{5}{*}{{Belief Merging}}
& GPT-5        & 6.88 & \bfseries 7.60 & 5.13 & 6.23 & 8.52 & 0.83 & -4.44 \\
& LLaMA-3B     & 6.89 & 6.88 & 6.42 & 6.62 & 7.94 & 0.84 & -4.18 \\
& LLaMA-8B     & \bfseries 7.21 & 7.23 & \bfseries 6.79 & \bfseries 6.96 & 8.31 & 0.84 & -4.15 \\
& Qwen3-4B     & 7.03 & 7.06 & 6.63 & 6.80 & 8.17 & 0.84 & -4.88 \\
& Qwen3-14B    & 7.13 & 7.18 & 6.71 & 6.93 & 8.20 & 0.84 & -4.16 \\

\cmidrule(lr){1-9}

\multirow{5}{*}{Belief Merging + Fusion}
& GPT-5        & 6.93 & 7.53 & 5.35 & 6.26 & 8.66 & 0.84 &  -4.12 \\
& LLaMA-3B     & 6.96 & 6.98 & 6.55 & 6.73 & 8.10 & 0.84 & -4.15 \\
& LLaMA-8B     &  7.15 &  7.18 &  6.77 &  6.93 &  8.31 & 0.84 & -4.14 \\
& Qwen3-4B     &  7.15 & 7.19 & 6.75 & 6.91 & 8.25 & 0.84 & -4.15 \\
& Qwen3-14B    & 6.95 & 6.98 & 6.53 & 6.72 & 8.02 &  0.85 & -4.15 \\

\bottomrule
\end{tabular}

\caption{
Evaluation results on the full dataset.
Bold indicates the best score per column.
}
\label{tab:full}
\end{table*}

\subsection{Criteria-Based Evaluation}
To evaluate summarisation quality along task-specific qualitative criteria that are not fully captured by reference-based metrics, we employ GEval \citep{liu2023gevalnlgevaluationusing}, an LLM-as-judge framework for structured evaluation. GEval prompts an LLM (OpenAI/gpt-4.1; \citealp{openai2024gpt4technicalreport}) to score generated summaries according to explicitly defined rubrics. Prompt templates and evaluation instructions are provided in appendix \ref{appendix:prompts}.

The evaluation criteria are chosen to reflect the core challenges of synthesis under disagreement, namely whether a summary preserves diverse viewpoints, maintains their polarity and prevalence, and remains grounded in the source reviews rather than collapsing conflict into spurious consensus. We configure GEval to assess summaries along the following dimensions: (1) \textbf{Opinion Coverage (Cov)}: inclusion of salient and contested opinions from the reviews;
    (2) \textbf{Polarity and Stance Fidelity (Pol)}: preservation of positive, negative, and mixed stances without distortion;
   (3) \textbf{Contrast Preservation (Ctr)}: explicit representation of disagreements across reviewers;
   (4) \textbf{Prevalence Calibration (Prev)}: accurate reflection of the relative prevalence of opinions;
   (5) \textbf{Groundedness (Grd)}: support of all claims by the source reviews.

Among these criteria, Cov, Pol, Ctr, and Prev assess disagreement-aware properties, while Grd evaluates factual groundedness independently of conflict. Each dimension is scored on a fixed Likert scale [1-10], and final GEval scores are obtained by averaging across dimensions.

\section{Results and Discussions}
Table~\ref{tab:full} reports the evaluation results on the full dataset. We analyse performance along disagreement-aware criteria, groundedness, and reference-based similarity. Overall, methods differ most strongly on disagreement-aware metrics, while reference-based metrics (BERT, BART) remain comparatively stable. This suggests that the primary challenge in this task lies in faithfully representing conflicting opinions, rather than producing fluent or semantically plausible summaries.

\paragraph{Effect of Explicit Aggregation} 
Across all models, belief merging yields consistently strong and stable performance on disagreement-aware criteria. Scores are tightly clustered across model families and sizes, indicating reduced sensitivity to architectural differences and parameter scale. This robustness suggests that explicitly resolving disagreement prior to generation shifts much of the aggregation burden away from the language model itself. Notably, this stability is achieved using simple, uniform synthesis prompts, without specialised aggregation-aware instructions. Even weaker models in this task, such as LLaMA-3B and LLaMA-8B, attain disagreement-aware scores that approach those of stronger models, such as GPT-5 and Qwen-14B, when guided by belief merging.
Comparing \emph{Belief Merging} with \emph{Belief Merging + Fusion-of-N} shows that applying more complex fusion-style prompting at generation time provides no systematic improvement once beliefs are explicitly aggregated. While small fluctuations are observed, belief merging alone already captures most of the achievable gains, indicating diminishing returns from additional prompt-level aggregation after belief-level resolution.

\paragraph{Prompt-Level Aggregation and Model Dependence}
Prompt-level aggregation methods, which perform aggregation during generation, exhibit substantially greater model dependence. Among these, Fusion-of-N performs strongly for capable models such as GPT-5 and Qwen-14B, achieving competitive disagreement-aware scores and high groundedness. This suggests that sufficiently strong models can internally reconcile conflicting inputs when guided by specialised fusion prompts.
However, this behaviour does not generalise across scales. For smaller LLaMA models, Fusion-of-N degrades sharply on coverage, contrast, and groundedness, indicating brittle aggregation when model capacity is limited. QA-based prompting, which also delegates aggregation to the model via intermediate questions, performs consistently below Fusion-of-N on disagreement-aware criteria, despite competitive reference-based scores. Together, these results suggest that prompt-level aggregation is inherently more sensitive to model capacity and less reliable across settings.
Overall, the comparison highlights a key trade-off: prompt-level aggregation can be effective for strong models but lacks robustness, whereas belief-level aggregation delivers stable disagreement-aware performance across models using simple prompting, reducing reliance on model scale or specialised fusion strategies.

\subsection{Manual Evaluation}
We conduct a targeted manual inspection on a subset of 50 movies selected from the test set to qualitatively validate the automatic evaluation. These movies were chosen because model outputs exhibited clear divergence on disagreement-aware GEval criteria, making them representative cases for assessing how different synthesis strategies handle conflicting opinions.

For each movie, we inspected summaries produced by (i) Fusion-of-N generation over reviews, (ii) belief merging with simple synthesis, and (iii) belief merging combined with Fusion-of-N–based realisation, across multiple model families. The inspection was performed by the authors and focused on concrete qualitative properties aligned with the evaluation criteria, including (i) coverage of salient aspects, (ii) preservation of opinion polarity, (iii) explicit representation of disagreement or mixed stances, and (iv) omission or suppression of minority viewpoints. This analysis is intended as a qualitative sanity check on the behaviours captured by GEval.

Under Fusion-of-N generation, behaviour varies substantially across models. Qwen models are more robust, typically covering a broader range of aspects and occasionally acknowledging split opinions, while GPT outputs are the most systematic, explicitly distinguishing majority and minority stances. LLaMA models frequently produce incomplete or unreliable summaries, including degenerate outputs and strong majority bias, often flattening disagreement and omitting contested or secondary aspects.
In contrast, belief merging markedly improves consistency across models: summaries show broader aspect coverage, clearer expression of contested opinions, and reduced suppression of minority views. These improvements are particularly pronounced for LLaMA models, whose outputs become more complete and balanced, closely aligning with their gains on disagreement-aware GEval metrics.
Applying Fusion-of-N at the synthesis stage yields only marginal additional qualitative gains beyond belief merging alone, primarily in occasional sharpening of contrast. Overall, the observed trends in manual inspection closely mirror the GEval results. An illustrative example is included in the Appendix \ref{app:example}.

\section{Conclusion}
Summarisation under disagreement poses challenges that are not primarily linguistic, but aggregative. When conflicting viewpoints are left to be reconciled during generation, performance becomes highly sensitive to model capacity and prompting strategies. In this work, we show that relocating aggregation outside generation, by explicitly merging beliefs prior to synthesis, yields more stable and faithful summaries across model families and scales. By grounding summarisation in belief-level aggregation, our approach preserves disagreement structure while allowing generation to focus on realisation rather than resolution. The results suggest that structured aggregation can substitute for increasingly complex prompting or larger models in settings where faithful synthesis under conflict is essential. More broadly, this work highlights the value of separating what is aggregated from how it is expressed, offering a principled direction for robust multi-document summarisation.

\section*{Limitations}
This work has limitations that suggest directions for future investigation.
First, a broader coverage across model architectures, scales, and training paradigms would be necessary to characterise the generality of the observed trends fully.
Second, while our results are consistent across the dataset considered, evaluation on a wider and more diverse range of opinion-heavy domains would strengthen claims about robustness and domain transferability.
Finally, our pipeline inherits constraints from the aspect extraction stage: some critic reviews do not explicitly mention all target aspects, which can limit downstream aggregation and lead to incomplete belief representations.
Improving aspect discovery and handling implicit or missing aspects, therefore, represents an important avenue for extending the applicability and reliability of the proposed approach.

\bibliography{custom}

\appendix

\section{Dataset Statistics and Implementation Details}
\paragraph{Dataset Statistics.} We present summary statistics associated with the evaluated dataset in Table \ref{tab:dataset}.
\label{appendix:dataset}
\begin{table*}[t]
\centering
\footnotesize
\setlength{\tabcolsep}{6pt} 
\begin{tabular}{lcc}
\toprule
\textbf{Statistic} & \textbf{Value}\\
\midrule
Nmuber of Movies & 914 \\
Reviews per Movie & 94 \\
Words per Review & 21  \\
Words per Meta-Review & 23 \\
\bottomrule
\end{tabular}
\caption{Rotten Tomato Dataset Statistics}
\label{tab:dataset}
\end{table*}

\paragraph{Implementation Details.}
 \label{appendix:taskdetails}
All experiments were conducted over approximately 13 days on a single NVIDIA A100. Inference with GPT-5  and GPT-4.1 via the OpenAI API cost approximately \$600 for all reviews.

\paragraph{Hyperparameters.}
The hyperparameters in Table ~\ref{tab:hyperparameters} were selected to encourage highly deterministic, concise outputs (temperature = 0) and to reduce redundancy. The maximum output token limit (200) was chosen to closely match the length distribution of gold critic meta-reviews and prevent differences in verbosity from confounding evaluation.

\begin{table*}[h!]
\centering
\footnotesize
\begin{tabular}{lccc}
\hline
\rule{0pt}{1.1em}
\textbf{Model} & \textbf{Role} & \textbf{Temperature} & \textbf{Max Tokens}\\
\hline
GPT-5 & Summariser & 0 & 200\\

Llama & Summariser &  0 & 200\\

Qwen & Summariser & 0 & 200 \\

GPT-4.1 & GEval & 0 & 2048\\

Qwen & Aspect Extraction & 0 & 512 \\  

Gemini & Data Subset Extraction & 0 & 1024\\
\hline
\end{tabular}
\caption{Hyperparameters used for different models in the experiment.}
\label{tab:hyperparameters}
\end{table*}

\section{Example Generation Across Models and Generation Strategies}
\label{app:example}
Here, in Table \ref{tab:qualitative_words_full} and \ref{tab:gold_words}, we provide a representative example of generated summaries across all models and generation strategies considered for a single movie: m/the\_words. Here we see that Fusion-of-N generation exposes strong architectural sensitivity, with smaller models failing to reliably synthesise content and larger models variably smoothing disagreement. Belief merging substantially mitigates this instability, yielding coherent, balanced summaries across all models that explicitly retain contested opinions.

\begin{table*}[t]
\centering
\scriptsize
\setlength{\tabcolsep}{5pt}
\begin{tabular}{p{1.6cm} p{2.2cm} p{8.8cm} p{2.8cm}}
\toprule
\textbf{Method} & \textbf{Model} & \textbf{Generated Summary} & \textbf{Notes} \\
\midrule

Fusion-of-N &
GPT &
The Words is an ambitious, multi-layered drama about literary plagiarism and the blur between fiction and reality, told through a story-within-a-story structure. Critics fault its glossy but shallow execution: the nested narratives are seen as confusing and needlessly convoluted, masking thin themes, blunt morals, and limited emotional depth. Heavy-handed narration and telegraphed twists contribute to a sense of pretension. While performances—particularly Jeremy Irons—and the film’s look and score receive some praise, the overall reception characterises it as a handsome but muddled and emotionally distant melodrama. &
Strong coverage; minority praise retained. \\

Fusion-of-N &
LLaMA-8B &
\textit{Degenerate output. Model produces code-generation instructions instead of a summary.} &
Task failure. \\

Fusion-of-N &
LLaMA-3B &
\textit{Degenerate output. The model produces partial NLP code giving instructions for producing a synthesis rather than the actual summary.} &
Task failure. \\

Fusion-of-N &
Qwen-14B &
The Words is a layered and ambitious film exploring authorship and blurred realities through a complex narrative structure. While praised for visual appeal and performances, it is widely criticised for convoluted storytelling, shallow literary treatment, and lack of emotional depth. Some appreciate its ambition, but overall it is viewed as a flawed and uneven literary drama that falls short of its potential. &
Disagreement compressed. \\

Fusion-of-N &
Qwen-4B &
Though the film features an ambitious, multi-layered narrative exploring plagiarism and artistic integrity, it is criticised for emotional distance, narrative incoherence, and superficial themes. Strong performances are acknowledged, but underdeveloped characters and exposition-heavy plotting undermine engagement. &
Broad but conventional. \\

\midrule

BM &
GPT &
Despite a visually polished presentation and committed performances, the film struggles to cohere. Its themes are engaged with clarity, but thin plotting, clumsy writing, and dragging pacing undermine the intended impact, resulting in a handsome yet ultimately unsuccessful work. &
Balanced, meta-review style. \\

BM &
LLaMA-3B &
Despite strong performances and striking visuals, the movie falls short due to an underwhelming and shallow plot, weak script coherence, and uneven pacing, ultimately delivering a disappointing experience. &
Stable, no degeneracy. \\

BM &
LLaMA-8B &
The film is a mixed bag: strong performances and visuals cannot compensate for an underdeveloped plot, lack of depth, and uneven pacing. Its thematic exploration is a minor highlight in an otherwise flawed film. &
Improved coverage. \\

BM &
Qwen-4B &
The film thoughtfully explores its themes and features vivid visuals and compelling performances, but a disjointed plot, weak narrative construction, and sluggish pacing prevent it from achieving cohesion or emotional resonance. &
Clear contrast. \\

BM &
Qwen-14B &
While visually appealing with strong performances and clearly conveyed themes, the film is undermined by an incoherent plot, poor structure, and inconsistent pacing, resulting in an unsuccessful overall experience. &
Consistent stance. \\

\midrule

BM + Fusion &
GPT &
Despite handsome visuals and strong performances, the film cannot overcome thin plotting, clumsy dialogue, and dragging pacing. Thematically potent ideas emerge intermittently, but the overall experience stumbles more than it succeeds. &
Stylistic refinement. \\

BM + Fusion &
LLaMA-3B &
Despite strong performances and visuals, the shallow plot, weak script, and uneven pacing prevent narrative engagement, resulting in a disappointing viewing experience. &
Minimal change vs. BM. \\

BM + Fusion &
LLaMA-8B &
The film falls short due to an underwhelming plot and lack of script coherence. While themes are explored effectively and performances are convincing, uneven pacing and narrative weakness dominate. &
Marginal gains only. \\

BM + Fusion &
Qwen-4B &
While the film’s visuals and performances are immersive and grounded, a poorly constructed script and sluggish pacing undermine narrative cohesion and emotional impact. &
Near-identical to BM. \\

BM + Fusion &
Qwen-14B &
Strong performances and cinematography are overshadowed by an underdeveloped plot, weak script, and inconsistent pacing, leaving an incomplete and unsatisfying experience. &
Diminishing returns. \\

\bottomrule
\end{tabular}
\caption{Qualitative comparison of summaries generated for a single movie across synthesis strategies and model families. Plain synthesis exhibits strong model-dependent instability, including degenerate failures, while belief merging yields coherent, disagreement-aware summaries across all models. Applying Fusion-of-N at realisation time provides only marginal additional gains.}
\label{tab:qualitative_words_full}
\end{table*}

\begin{table*}[t]
\centering
\small
\begin{tabular}{p{12.5cm}}
\toprule
\textbf{Critics' Consensus (Reference)} \\
\midrule
Neither as clever nor as interesting as it appears to think it is, \textit{The Words} maroons its talented stars in an overly complex, dramatically inert literary thriller that is ultimately a poor substitute for a good book. \\
\bottomrule
\end{tabular}
\caption{Reference critics’ consensus for the example movie, shown for contextual grounding only. This text is not used for training or evaluation.}
\label{tab:gold_words}
\end{table*}

\section{LLM Prompts}
\label{appendix:prompts}
Here we present the prompts used within our experiments. \emph{FUSION\_ONLY\_PROMPT} represents the baseline adaptation of Fusion-of-N \citep{khairi2025makingtakingbestn}, \emph{WORLD\_CONVERSION\_PROMPT} and \emph{BELIEF\_MERGING\_PROMPT} represent the prompts used with our proposed pipeline, belief merging, both for converting quantitative belief bases to natural language hypotheses and for generating summaries. The remaining prompts, \emph{BM\_FUSION\_PROMPT} and \emph{GEVAL\_PROMPT} represent prompts used with the other merged belief base summarisation strategy (Belief Merging + Fusion-of-N) and used to generate the reported GEVAL scores. We use one-shot prompting across all our experiments.

\begin{lstlisting}[style=promptstyle]
FUSION_ONLY_PROMPT = 

You are given a set of input reviews. Your task is to generate a **synthesized summary** of these reviews.

Instructions:

1. **Capture all aspects:** Read all the input reviews carefully and ensure that your summary reflects the full range of opinions, points, and details mentioned. Do not omit any significant aspect of the reviews.

2. **Be concise and coherent:** Present the information in a well-structured, readable summary.

3. **Maintain neutrality:** Do not inject your own opinion; the summary should reflect the content and tone of the input reviews faithfully.

One-shot example:

Input:
[
    'A fantasy adventure that fuses Greek mythology to contemporary American places and values. Anyone around 15 (give or take a couple of years) will thrill to the visual spectacle',
    'Uma Thurman as Medusa, the gorgon with a coiffure of writhing snakes and stone-inducing hypnotic gaze is one of the highlights of this bewitching fantasy',
    'With a top-notch cast and dazzling special effects, this will tide the teens over until the next Harry Potter instalment.',
    "Whether audiences will get behind The Lightning Thief is hard to predict. Overall, it's an entertaining introduction to a promising new world -- but will the consuming shadow of Potter be too big to break free of?",
    "What's really lacking in The Lightning Thief is a genuine sense of wonder, the same thing that brings viewers back to Hogwarts over and over again.",
    "It's more a list of ingredients than a movie-magic potion to enjoy from start to finish.",
    "Harry Potter knockoffs don't come more transparent and slapdash than this wannabe-franchise jumpstarter directed by Chris Columbus.",
    "Percy Jackson isn't a great movie, but it's a good one, trotting out kernels of Greek mythology like so many Disney Channel references.",
    'Fun, brisk and imaginative'
]

Output:
'Though it may seem like just another Harry Potter knockoff, Percy Jackson benefits from a strong supporting cast, a speedy plot, and plenty of fun with Greek mythology.'

Input format: [<List of Input Reviews>]

Output format:
A single coherent summary that synthesizes all input reviews following the instructions above.

\end{lstlisting}

\begin{lstlisting}[style=promptstyle]
WORLD_CONVERSION_PROMPT = 
You are given a binary merged-world interpretation (derived through belief merging) representing collective critic opinions across seven movie aspects.

Movie Aspects:
1. Plot
2. Themes
3. Script
4. Performances
5. Visuals
6. Pacing
7. Overall Success

Each aspect has a binary value:
- "1" = positive or favorable
- "0" = negative or unfavorable

Your task:
Generate a natural language hypothesis or interpretation for each aspect, clearly expressing what the merged-world judgment implies about the movie. Each hypothesis should be short, fluent, and sound like a critic's natural statement (not a label).

Output format (strictly follow this JSON structure):
{{
  "Plot Hypothesis": "...",
  "Themes Hypothesis": "...",
  "Script Hypothesis": "...",
  "Performances Hypothesis": "...",
  "Visuals Hypothesis": "...",
  "Pacing Hypothesis": "...",
  "Overall Success Hypothesis": "..."
}}
Example
Input World:
{{
  "Plot": "1",
  "Themes": "1",
  "Script": "0",
  "Performances": "1",
  "Visuals": "1",
  "Pacing": "0",
  "Overall Success": "1"
}}

Expected Output:
{{"Plot Hypothesis": "The movie plot is engaging and well developed.",
  "Themes Hypothesis": "The film explores its themes effectively.",
  "Script Hypothesis": "The script lacks depth and coherence.",
  "Performances Hypothesis": "The performances are strong and convincing.",
  "Visuals Hypothesis": "The visuals are striking and memorable.",
  "Pacing Hypothesis": "The pacing feels uneven and occasionally slow.",
  "Overall Success Hypothesis": "Overall, the film delivers a successful and enjoyable experience."
}}
Now, generate the corresponding natural language hypotheses for the following merged-world interpretation:
{text}
\end{lstlisting}

\begin{lstlisting}[style=promptstyle]
BM_FUSION_PROMPT = 
You are given natural language interpretations derived from a merged-world representation produced through belief merging on a movie review dataset.

Your role:
Act as a **fusor and synthesizer**. Combine all unique insights across the provided aspect-level interpretations into a single, coherent **meta-review**. This meta-review should:
- Seamlessly integrate the key points from all aspects.
- Reflect the general tone and balance of the interpretations.
- Read naturally, like a critic's concise, well-written summary.
- Avoid repetition or listing; aim for fluent synthesis.

Output format (strictly follow this JSON structure):
{{
  "Proposed Meta-Review": "..."
}}
---
### Example
**Input Interpretations:**
{{
  "Plot Hypothesis": "The movie's plot is engaging and well developed.",
  "Themes Hypothesis": "The film explores its themes effectively.",
  "Script Hypothesis": "The script lacks depth and coherence.",
  "Performances Hypothesis": "The performances are strong and convincing.",
  "Visuals Hypothesis": "The visuals are striking and memorable.",
  "Pacing Hypothesis": "The pacing feels uneven and occasionally slow.",
  "Overall Success Hypothesis": "Overall, the film delivers a successful and enjoyable experience."
}}
**Expected Output:**
{{
  "Proposed Meta-Review": "The film succeeds as an engaging and visually captivating experience, supported by strong and convincing performances. Its well-developed plot and thoughtfully explored themes create a solid emotional foundation, even though the script occasionally lacks depth and coherence. Despite moments of uneven pacing, the movie's overall execution feels confident and enjoyable, resulting in a work that resonates both narratively and aesthetically."
}}
---
Now, generate the proposed meta-review for the following interpretations:
{text}

\end{lstlisting}

\begin{lstlisting}[style=promptstyle]
BELIEF_MERGING_PROMPT = 
You are given natural-language interpretations derived from multiple hypotheses and analyses of a movie's reviews.

Your task is to generate an accurate summarised meta-review that reflects the combined meaning of these interpretations.

Requirements for the summary:
- Capture all key points conveyed across the interpretations.
- Maintain a balanced, neutral tone.
- Represent differing perspectives fairly (positive, negative, mixed).
- Read fluently like a concise professional critic's summary.
- Avoid repetition, bullet-style listing, or enumerating points.
- Do not introduce opinions not grounded in the input.

Your output must strictly follow the JSON format below:

{{
  "Proposed Meta-Review": "..."
}}

---

### Example

**Input Interpretations:**
{{
  "Plot Hypothesis": "The movie's plot is engaging and well developed.",
  "Themes Hypothesis": "The film explores its themes effectively.",
  "Script Hypothesis": "The script lacks depth and coherence.",
  "Performances Hypothesis": "The performances are strong and convincing.",
  "Visuals Hypothesis": "The visuals are striking and memorable.",
  "Pacing Hypothesis": "The pacing feels uneven and occasionally slow.",
  "Overall Success Hypothesis": "Overall, the film delivers a successful and enjoyable experience."
}}

**Expected Output:**
{{
  "Proposed Meta-Review": "The film succeeds as an engaging and visually captivating experience, supported by strong and convincing performances. Its well-developed plot and thoughtfully explored themes create a solid emotional foundation, even though the script occasionally lacks depth and coherence. Despite moments of uneven pacing, the movie's overall execution feels confident and enjoyable, resulting in a work that resonates both narratively and aesthetically."
}}

---

Now, generate the proposed meta-review for the following interpretations:

{text}

\end{lstlisting}

\begin{lstlisting}[style=promptstyle]

GEVAL_PROMPT = 

  You are an expert evaluator of summaries of multiple movie reviews that contain conflicting opinions.

  Assume the summary is fluent and coherent.

  Your task is to evaluate whether the summary faithfully represents the content of the source reviews.

  Focus only on opinion coverage, polarity fidelity, contrast preservation, prevalence calibration,

  and groundedness. Do not judge style or writing quality.

  Evaluate the generated summary below.

  source_reviews: {{reviews}}

  generated_summary: {{summary_text}}

  aspect_schema: Focus on these aspects:
  - Plot, Themes, Script, Performances, Visuals, Pacing, Overall Success

  evaluation_steps:

    step_1_extract_opinion_units:

      description:

        Extract opinion units from the source reviews.

        Each opinion unit must include:

          - aspect

          - polarity (positive, negative, mixed or contested, neutral)

          - intensity (mild, moderate, strong)

          - short rationale phrase grounded in the source

          - prevalence (minority, split, majority, unclear)

        Extract between eight and twenty distinct opinion units.

        Merge duplicates but preserve meaningful differences.


    step_2_check_coverage:

      description:

        For each extracted opinion unit, determine whether it is:

          - covered accurately in the summary

          - covered but distorted (wrong polarity, aspect, or implied prevalence)

          - missing

        Also identify any opinions stated in the summary that are not supported by the source reviews.
    step_3_score_dimensions:

      description:

        Score each dimension independently from zero to ten using whole numbers only.

      dimensions:

        opinion_coverage:

          question: How completely does the summary include the important opinion units, especially those that are contested or central?


        polarity_and_stance_fidelity:

          question: Are positive, negative, and mixed opinions preserved without reversal, dilution, or flattening?


        contrast_preservation:

          question: Does the summary clearly express disagreements, such as some reviewers praising an aspect while others criticize it?

        prevalence_calibration:

          question: Does the summary avoid implying consensus when opinions are split, or exaggerating minority views?



        groundedness:

          question: Are all opinions and claims in the summary supported by the source reviews?



    step_4_output:

      format: Provide the output with the following fields:

        opinion_units:

          [<list of extracted opinion units>]


        coverage_assessment:

          [<list of opinion units with coverage labels>]


        dimension_scores:

          opinion_coverage: 0-10

          polarity_and_stance_fidelity: 0-10

          contrast_preservation: 0-10

          prevalence_calibration: 0-10

          groundedness: 0-10


        unsupported_summary_opinions:

          [<list>]


        evaluator_notes:

          short explanation of major errors and omissions
\end{lstlisting}

\section{Topic Modelling for Aspect Extraction}
Here we present and highlight the Topic Modelling done to determine aspects to extract from each Movie review.
\begin{figure}[ht]  
    \centering
    \includegraphics[width=0.5\textwidth]{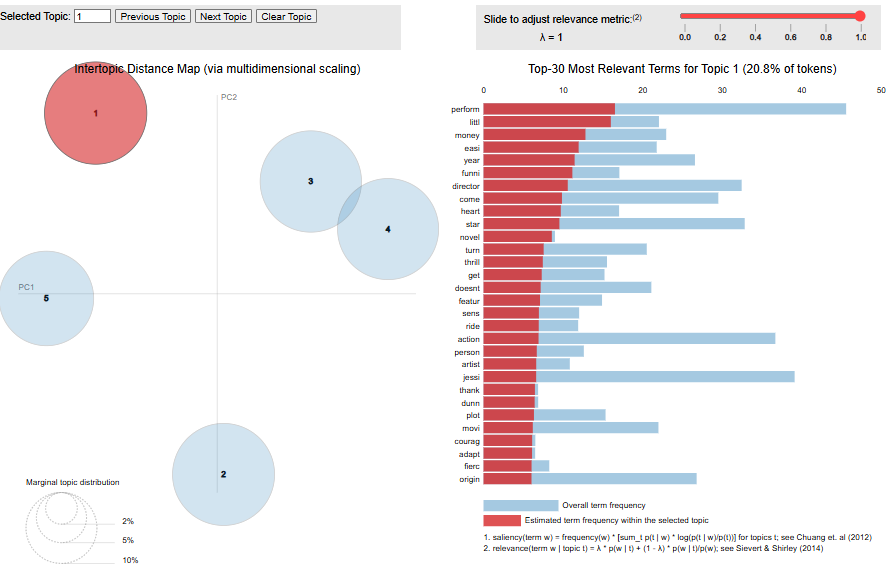}  
    \caption{Topic Modelling to determine aspects to extract from Reviews for further downstream processing. Highlighting the results of Topic 1}
    \label{fig:topic_model}  
\end{figure}

\section{Belief Merging Procedure for Critic Evaluations}
\label{app:belief-merging}

This appendix details the belief merging procedure used to aggregate multiple critics' evaluations of a movie into a single merged belief base \(K^{\ast}\). The procedure follows a distance-based approach with probabilistic aggregation.

\subsection{Aspects and Interpretations}

Let 
\[
\mathcal{A} = \{a_1, a_2, \dots, a_7\}
\]
denote the set of evaluated aspects:
\begin{quote}
Plot, Themes, Script, Performances, Visuals, Pacing, Overall success
\end{quote}

Each aspect \(a_j\) is represented as a binary propositional variable, with \(1\) indicating a positive assessment and \(0\) indicating a negative assessment. A \emph{world} (or interpretation) \(w \in \mathcal{W}\) is a complete assignment of 0/1 values to all aspects:
\[
\mathcal{W} = \{0,1\}^{|\mathcal{A}|}.
\]

\subsection{Critic Profiles}

For each movie \(m\), let
\[
C_m = \{c_1, \dots, c_n\}
\]
be the set of critics providing evaluations. Each critic \(c \in C_m\) gives a graded score
\[
c(a_j) \in [0,1]
\]
for aspects \(a_j\). Missing or undefined values are ignored during computation.

\subsection{Distance Between a World and a Critic}

The disagreement between a world \(w\) and a critic \(c\) is measured using an L1 (Manhattan) distance:
\[
d(w,c) = \sum_{j=1}^{|\mathcal{A}|} \delta(w_j, c(a_j)),
\]
where
\[
\delta(w_j, c(a_j)) =
\begin{cases}
1 - c(a_j), & \text{if } w_j = 1, \\
c(a_j), & \text{if } w_j = 0,
\end{cases}
\]
and undefined scores are ignored.  

If all critic scores are binary (0 or 1), this reduces to the Hamming distance between \(w\) and the critic's evaluation.

\subsection{Distance-Based Belief Merging}

For each movie \(m\), the total distance of a world \(w\) to all critics is
\[
D_m(w) = \sum_{c \in C_m} d(w,c).
\]

The set of optimal worlds minimising the total distance is
\[
\mathcal{W}^\ast_m = \arg\min_{w \in \mathcal{W}} D_m(w).
\]

These worlds represent the interpretations that are collectively closest to all critic evaluations.

\subsection{Probabilistic Merged Belief Base \(K^{\ast}\)}

When multiple worlds minimise \(D_m(w)\), aggregation is performed by averaging across the selected worlds. For each aspect \(a_j\), the merged belief base assigns
\[
K^{\ast}_m(a_j) = \frac{1}{|\mathcal{W}^\ast_m|} \sum_{w \in \mathcal{W}^\ast_m} \mathbf{1}[w_j = 1],
\]
where \(\mathbf{1}[\cdot]\) is the indicator function.  

Thus, \(K^\ast_m(a_j) \in [0,1]\) can be interpreted as the probability that aspect \(a_j\) is judged positively in a probabilistic compromise across critics.

\subsection{Summary of Outputs}
For each movie, the procedure produces:
\begin{itemize}
    \item The set of optimal worlds \(\mathcal{W}^\ast_m\) with minimal total distance.
    \item The merged belief base \(K^{\ast}_m\), giving a probabilistic assessment for each aspect.
\end{itemize}

This approach captures both the logical structure of possible collective judgments and the probabilistic compromise when multiple interpretations equally minimise disagreement.

\end{document}